\documentclass{article} 
\usepackage{iclr2023_conference_tinypaper,times}

\usepackage{amsmath,amsfonts,bm}

\def\eqref#1{equation~\ref{#1}}

\def\1{\bm{1}}

\DeclareMathAlphabet{\mathsfit}{\encodingdefault}{\sfdefault}{m}{sl}
\SetMathAlphabet{\mathsfit}{bold}{\encodingdefault}{\sfdefault}{bx}{n}

\usepackage{hyperref}
\usepackage{graphicx}
\usepackage{url}

\title{Enhancing Language Models for Financial Relation Extraction with Named Entities and Part-of-Speech}

\author{Menglin Li \& Kwan Hui Lim \\
Singapore University of Technology and Design \\
\texttt{menglin\_li@mymail.sutd.edu.sg, kwanhui\_lim@sutd.edu.sg}
}

\iclrfinalcopy 
\begin{document}

\maketitle

\begin{abstract}
The Financial Relation Extraction (FinRE) task involves identifying the entities and their relation, given a piece of financial statement/text. To solve this FinRE problem, we propose a simple but effective strategy that improves the performance of pre-trained language models by augmenting them with Named Entity Recognition (NER) and Part-Of-Speech (POS), as well as different approaches to combine these information. Experiments on a financial relations dataset show promising results and highlights the benefits of incorporating NER and POS in existing models.
Our dataset and codes are available at \url{https://github.com/kwanhui/FinRelExtract}.
\end{abstract}

\section{Introduction}

Financial Relation Extraction (FinRE) is an important task for the financial domain, which involves the identification of key entities in a piece of text and the relation between these entities~\cite{kaur2023refind,hillebrand2022kpi}. Figure~\ref{FinREproblem} shows an illustration of this problem. The more generic task of Relation Extraction also has great significance for the general field of Natural Language Processing~\cite{wang2022deep,nayak2021deep}. The Bidirectional Encoder Representations from Transformers (BERT)~\cite{devlin2018bert} and its variants have been proposed and shown good results for FinRE and related tasks~\cite{hillebrand2022kpi,qiu2023simple,vardhan2023imetre}. For example, KPI-BERT~\cite{hillebrand2022kpi} utilizes BERT with a RNN-based pooling mechanism for FinRE in the context of Key Performance Indicators and their relations. Others like~\cite{qiu2023simple} build upon R-BERT~\cite{wu2019enriching} by leveraging both sentence-level and entity-level information for the FinRE task, while~\cite{vardhan2023imetre} utilize a similar approach but with XLNET~\cite{yang2019xlnet} as the backbone architecture, with additional entity-type markers in the text. Building upon these works, we leverage on pre-trained language models and propose various strategies for augmenting additional information into the standard financial text.

In this paper, we make a few contributions: (i) We propose a simple but effective strategy of using Named Entity Recognition (NER) and Part Of Speech (POS) to improve the performance of pre-trained language models on the FinRE task; (ii) We study the effects of considering NER and POS in these models, as well as different ways of combining this information; and (iii) We perform a series of experiments and ablation study that shows the strong performance of our proposed method.

\section{Methods}
\label{sectMethods}

{\noindent\bf Intuition and Research Questions.} Our intuition is that pre-trained language models can be enhanced with additional NER and POS information, as the FinRE task involves identifying the relation between two entities. For example in the ``Google:acquired:Fitbit'' relation, NER is useful for identifying entities (e.g., Fitbit and Google) while POS is useful for identifying verbs that are potentially relations (e.g., acquired). As such, we consider the following two Research Questions (RQs) when developing our proposed model, namely: (RQ1) How can we utilize NER and POS to improve pre-trained language models for FinRE? and (RQ2) What is the best strategy for combining NER and POS  with its original text content?

{\noindent\bf Proposed Model.} The foundation of our model is a RoBERTa~\cite{liu2019roberta} backbone, with various strategies for incorporating NER and POS, alongside the original financial text. Given a financial text $T = (t_1, t_2, ... t_n)$ with $n$ tokens, we first extract the corresponding list of NER tokens $E = (e_1, e_2, ... e_n)$ and POS tokens $P = (p_1, p_2, ... p_n)$. For combining text and NER, we use the Text $T$ replaced by NER tokens $E$ that are not ``Others'', resulting in $TrN = (t_1, e_2, ..., e_{17}, ... t_n)$ where $e_2$ and $e_17$ are NER tokens among the original text. This text with replacement by NER sequence $TrN$ is then concatenated with the POS sequence $P$, now denoted $TrNP$, before being fed into a RoBERTa backbone. Figure~\ref{modelArchitecture} shows our proposed model architecture. In our later ablation study, we also study the effect of different combinations of these text, NER and POS components.

\section{Experiments and Results}
\label{sectResults}

\textbf{\noindent Dataset and Experiment Setup.} We experimented using the REFinD dataset~\cite{kaur2023refind}, which comprises 28,676 annotated instances of 22 different financial relations between eight entity types based on US financial filings. REFinD contains 20,070, 4,306 and 4,300 instances for the training, development and test sets, respectively. We fine-tune our proposed model and the baselines (described later) with the AdamW optimizer using a learning rate of $lr2e-05$, decay of 0.1, dropout of 0,1, batch size of 32 and for 5 epochs. Early stopping is employed and the best model selected from the development set. Micro-F1 and Macro-F1 scores are reported on the test set.

\textbf{\noindent Results: Models and Baselines.} 
In addition to our proposed model, we evaluated it against various other baseline models with the Text (T) input, i.e., the text represents the financial statement. These other models include:
SpanBERT~\cite{joshi2020spanbert}, 
FinBERT~\cite{yang2020finbert},
BERT~\cite{devlin2018bert},
XLM-RoBERTa~\cite{conneau2020unsupervised},
DistilBERT~\cite{sanh2019distilbert} and
ALBERT~\cite{lan2019albert},
and the results are shown in Table~\ref{expResults}. The results show the superior performance of our proposed model against the baselines, with absolute improvements of 0.1144 in Micro-F1 and 0.2134 in Macro-F1, against the best performing baseline models.
In addition, we conducted further experiments to compare the same baseline models using the same input, i.e., Text with replacement of NER + POS ($TrNP$), as shown in Table~\ref{expResults2}. Despite using the same input, the results show that our proposed model still offers the best performance compared to other baseline models, with the financial-related FinBERT being a close second in terms of both Micro-F1 and Macro-F1 scores. An additional ablation study is also presented and discussed in Appendix~\ref{sectAblationStudy}.
\vspace{-3mm}
\begin{table} [ht]
\def\arraystretch{1.1}
\centering
\parbox{.48\linewidth}{
\caption{Experimental Results of Proposed Model against Baselines using Text (T).}
\label{expResults}
\centering
\begin{tabular}{c c c} \hline
{Model} & {Micro-F1}	&	{Macro-F1}	 \\ \hline
SpanBERT	&	0.6556	& 0.2505	\\
FinBERT	&	0.6477	& 0.3806	\\
BERT	&	\textit{0.6577}	& 0.3311	\\
XLM-RoBERTa	&	0.6407	& \textit{0.3373}	\\
ALBERT	&	\textit{0.6577}	& 0.2910	\\
DistilBERT	&	0.6463	& 0.3281	\\
Proposed Model	&	\textbf{0.7721}	&	\textbf{0.5507}	\\ \hline
\end{tabular}
}
\hspace{3mm}
\parbox{.48\linewidth}{
\caption{Experimental Results of Proposed Model against Baselines using Text with replacement of NER + POS (TrNP).}
\label{expResults2}
\centering
\begin{tabular}{c c c} \hline
{Model} & {Micro-F1}	&	{Macro-F1}	 \\ \hline
SpanBERT	&	0.7572	&	0.3504	\\
FinBERT	&	0.7609	&	\textit{0.5341}	\\
BERT	&	0.7702	&	0.4612	\\
XLM-RoBERTa	&	0.7693	&	0.4597	\\
ALBERT	&	0.7447	&	0.4622	\\
DistilBERT	&	\textit{0.7712}	&	0.4800	\\
Proposed Model	&	\textbf{0.7721}	&	\textbf{0.5507}	\\ \hline
\end{tabular}
}
\end{table}
\vspace{-3mm}

\begin{table} [h!]
\def\arraystretch{1.1}
\centering
\end{table}

\section{Conclusion and Future Work}
We propose a simple but effective strategy to enhance pre-trained language model for the FinRE task, by replacing financial text with NER tokens, then concatenating with its corresponding POS tokens. Experimental results on the REFinD dataset show the promising performance of this approach. Our study also shows that this NER replacement strategy performs better than simply concatenating the NER tokens, and NER has a bigger role in improving performance than POS for such models. Future directions include a more in-depth study on the effects of different types of NER and POS tokens, as well as adaptation to other domains such as for geospatial predictions and recommendations~\cite{li2023transformer,halder2023survey}.

\newpage
\section{URM Statement}
The authors acknowledge that at least one key author of this work meets the URM criteria of ICLR 2024 Tiny Papers Track.

\bibliography{FinRelExt}
\bibliographystyle{iclr2023_conference_tinypaper}

\appendix
\section{Appendix}

\subsection{Additional Details about the Financial Relation Extraction (FinRE) Problem}

\begin{figure}[ht]
    \centering
    \includegraphics[width=\linewidth]{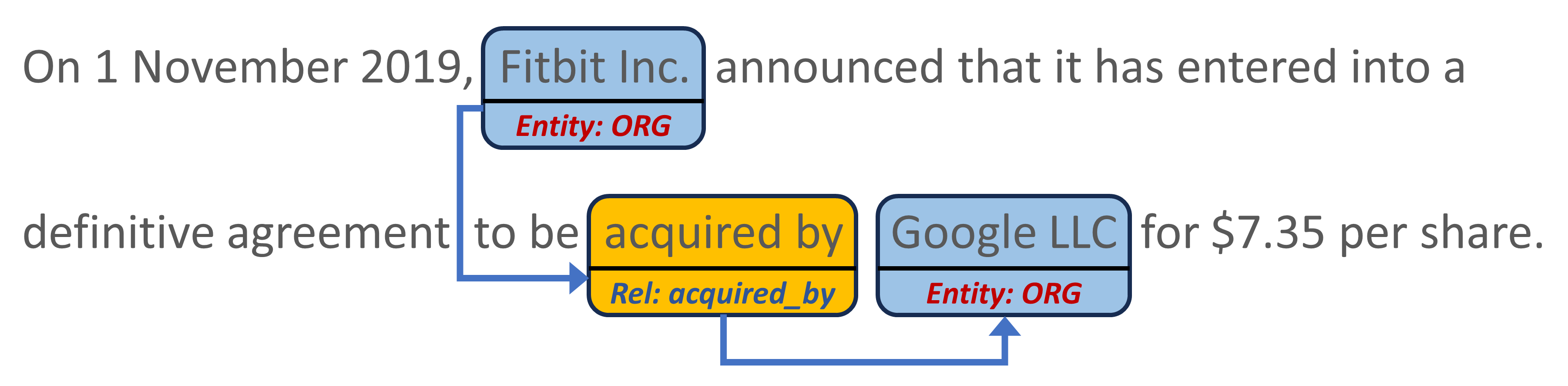}
    \caption{An example of the Financial Relation Extraction (FinRE) problem, with the relation \textit{org:org:acquired\_by} extracted from this financial statement.}
    \label{FinREproblem}
\end{figure}

The Financial Relation Extraction (FinRE) problem involves identifying the relation between entities in a financial statement, i.e., a piece of text relating to finance matters. While the eventual ground truth label is in the form of a financial relation tag, this task also involves the additional sub-tasks of detecting the key entity types and identifying the relationship between them. Figure~\ref{FinREproblem} shows an example of this problem, where the financial relation label is ``Google:acquired:Fitbit'', which is in turn based on the entities ``Google'' and ``Fitbit'' connected by a ``acquired'' relation.

For our experiments, we use the REFinD dataset~\cite{kaur2023refind}, comprising 28,676 annotated instances with 22 different financial relations. There are also other relevant datasets for the FinRE tasks, such as TACRED~\cite{zhang2017position}, KBP37~\cite{zhang2015relation},  FinRED~\cite{sharma2022finred} and CorpusFR~\cite{jabbari2020french}. Among these dataset, TACRED is the largest with 119,474 instances and 42 relations and CorpusFR the smallest with 1,754 instances and 20 relations. REFinD is the second largest among these datasets but poses a distinct advantage over the largest TACRED in that the REFinD only comprises 45.5\% instances of the ``no relation'' type, while TACRED has a larger proportion at 79.5\%. Thus, our choice of REFinD for our experiments.

\subsection{Additional Details about Proposed Model}

\begin{figure}[h!]
    \centering
    \includegraphics[width=\linewidth]{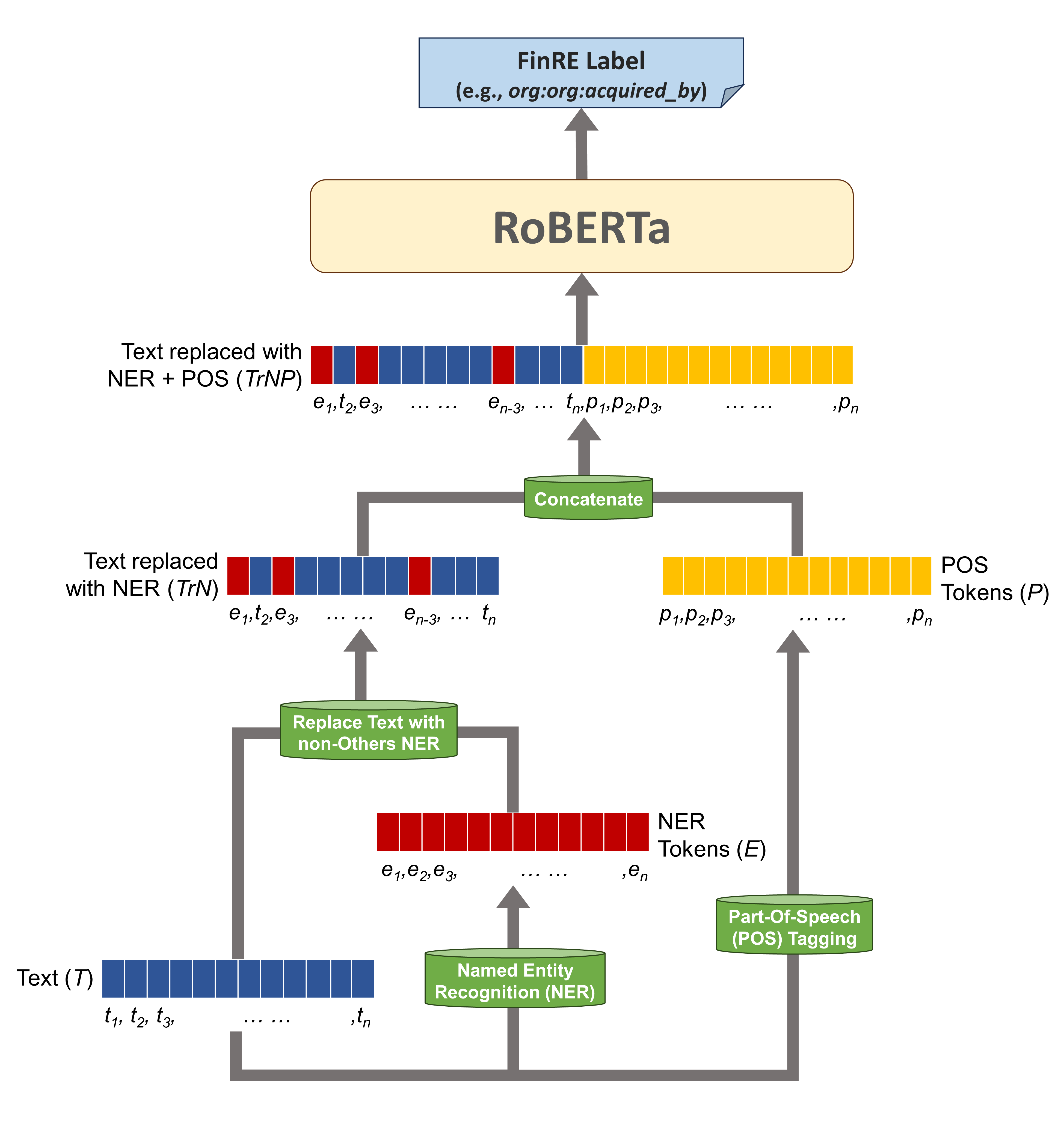}
    \caption{Architecture of Our Proposed Model for the FinRE Task.}
    \label{modelArchitecture}
\end{figure}

Figure~\ref{modelArchitecture} provides an illustration of our proposed model architecture. As mentioned in Section~\ref{sectMethods}, the foundation of our model is a RoBERTa~\cite{liu2019roberta} backbone and we build upon the roberta-base implementation on Hugging Face (https://huggingface.co/roberta-base) for this purpose. Thereafter, we propose and explore various strategies for incorporating NER and POS information to enhance the performance in the FinRE task. We selected RoBERTa over BERT and other variants due to the overall stronger performance of the former over a range of text-related tasks, via the implementation of a modified pre-training methodology. While the training time for RoBERTa might be longer, we do not consider this a significant issue as we are using RoBERTa as a pre-trained language model and fine-tuning it using our NER and POS augmented approach.

Our input is in the form of a financial text $T = (t_1, t_2, ... t_n)$ with $n$ tokens. Using this financial text $T$ input, we utilize the spaCy library (https://spacy.io/) to extract the list of NER tokens $E = (e_1, e_2, ... e_n)$ and POS tokens $P = (p_1, p_2, ... p_n)$, as indicated by the green boxes labelled ``Named Entity Recognition (NER)'' and ``Part-Of-Speech (POS) Tagging'' in Figure~\ref{modelArchitecture}. For both NER and POS, we use the tag/token type and not the actual entity names, i.e., ``[ORG]'' and not ``Microsoft''. This sequence of NER tokens $E$ (e.g., ``[ORG]'') are then used to replace their text counterparts $T$, resulting in a sequence $TrN$ with the same length. For example, if $e_2$ and $e_17$ are the detected NER tokens, then the resulting sequence will be $TrN = (t_1, e_2, ..., e_{17}, ... t_n)$. Following which, we then concatenate this text with replacement by NER sequence $TrN$ with the POS sequence $P$, separated by the [SEP] [CLS] tokens typically used in BERT and its variants. Finally, this concatenated sequence $TrNP$ is provided as input into a RoBERTa model and trained using the provided FinRE labels.

\subsection{Additional Ablation Study and Discussion}
\label{sectAblationStudy}

As part of our ablation study, we performed additional experiments to evaluate the different components of Text (T), NER (N), POS (P), Text with replacement by NER (TrN). We evaluated six combinations of these components as shown in Table~\ref{ablationStudy}, e.g., TNP refers to RoBERTa using the Text, NER and POS components as a concatenated input. These results show the effectiveness of including both NER and POS, particularly our strategy of replacing text with NER tokens (TrNP), compared to simply concatenating NER to the original text (TNP).

\begin{table} [ht]
\def\arraystretch{1.1}
\centering
\caption{Ablation Study (T = Text, N = NER, P~=~POS, TrN = Text with replacement of NER).}
\label{ablationStudy}
\centering
\begin{tabular}{c c c} \hline
{Component} & {Micro-F1}	&	{Macro-F1}	 \\ \hline
T	&	0.6507	&	0.3231	\\
TN	&	0.7265	&	0.5342	\\
TP	&	0.6605	&	0.3546	\\
TNP	&	0.7379	&	0.4702	\\
TrN	&	\textit{0.7640}	&	\textbf{0.5559}	\\
TrNP	&	\textbf{0.7721}	&	\textit{0.5507}	\\ \hline
\end{tabular}
\end{table}

In the previous paragraph, we have seen that Table~\ref{ablationStudy} highlights the effectiveness of replacing the financial text with NER tokens and concatenating the POS token sequence. We further note that both strategies of replacement by NER tokens and concatenating POS token sequence are also effective individually in improving performance, although NER tokens have a larger effect than POS tokens. This trend is evident in how TN and TrN have a larger improvement over T in terms of both Micro-F1 and Macro-F1, as compared to TP that has a much smaller improvement over T. 

\end{document}